\documentclass{article}
\usepackage{booktabs}
\usepackage{multirow}
\usepackage{array}
\usepackage{amsmath}
\usepackage{amssymb} 
\usepackage{graphicx,svg}
\usepackage{float}

\usepackage{graphicx}
\usepackage{subcaption}
\usepackage{float}
\usepackage{caption}
\usepackage[preprint]{neurips_2026}

\usepackage{booktabs}
\usepackage{multirow}
\usepackage{array}
\usepackage{amsmath}
\usepackage{amssymb} 
\usepackage{graphicx,svg}
\usepackage{float}
\usepackage{enumitem}
\usepackage{graphicx}
\usepackage{subcaption}
\usepackage{float}
\usepackage{caption}

\usepackage[utf8]{inputenc} 
\usepackage[T1]{fontenc}    
\usepackage{hyperref}       
\usepackage{url}            
\usepackage{booktabs}       
\usepackage{amsfonts, amsthm}       
\usepackage{nicefrac}       
\usepackage{microtype}      
\usepackage{xcolor}         
\usepackage{algorithm, algpseudocode}

\newtheorem{proposition}{Proposition}

\title{Learn What’s Left, Not What’s Mastered: Saturation Aware Advantage Reweighting for Multi-Reward Policy Optimization}

\usepackage{cleveref}

\author{%
  \textbf{Yixuan Wang}$^{\spadesuit *}$\quad
  \textbf{Yifei Chen}$^{\heartsuit *}$\quad
  \textbf{Haichao Zhang}$^{\clubsuit}$\quad
  \textbf{Haozheng Luo}$^{\diamondsuit}$\quad
  \textbf{Xander Wu}$^{\bigstar}$\\
  \textbf{Jie Ni}$^{\blacklozenge}$\quad
  \textbf{Yun Fu}$^{\clubsuit}$\quad
  \textbf{Nuno Vasconcelos}$^{\heartsuit}$\quad
  \textbf{Yijiang Li}$^{\heartsuit \dagger}$
  \\
  {\normalfont
    $^{\spadesuit}$University of Florida \quad
    $^{\heartsuit}$UC San Diego \quad
    $^{\clubsuit}$Northeastern University}\\
  {\normalfont
    $^{\diamondsuit}$Northwestern University \quad
    $^{\bigstar}$Stanford University, Zillion Network \quad
    $^{\blacklozenge}$Universität Innsbruck}\\
  {\normalfont\footnotesize $^{*}$Equal contribution. \quad $^{\dagger}$Corresponding author \& project lead.}\\
  {\normalfont\texttt{yijiangli@ucsd.edu}}
}

\begin{document}

\maketitle
\begin{abstract}
  Reinforcement learning (RL) with group-relative advantages has become the de facto standard for post-training language model reasoners.
  However, when optimizing multiple reward objectives, existing methods typically scalarize the reward vector with a fixed weighted sum before group-wise standardization. We show that this design leads to two fundamental problems: rollouts with distinct reward profiles can receive identical advantages, and all objectives are optimized with fixed relative weights regardless of their current level of saturation. Consequently, an objective whose rewards are already near their upper bound can retain substantial influence when its rewards still vary within rollout groups.
   We introduce \textbf{Saturation Aware Advantage Reweighting for Multi-Reward Policy Optimization} (SA-MRPO), which standardizes each reward objective independently and adaptively discounts its contribution according to a batch-level estimate of objective saturation. This changes the relative contribution of each objective according to its observed reward headroom. We further derive an exact condition under which saturation aware reweighting reverses the sign of a rollout's aggregate advantage.
   %
   Across mathematical reasoning with two- and three-objective reward combinations, SA-MRPO improves the harder correctness objective over GDPO in 12 of 15 benchmark comparisons, with gains of up to $5\%$ on AIME24.
   On adaptive reasoning it improves accuracy on all five benchmarks, by $3.8\%$ on average and up to $9.2 \%$ on AMC23, and on coding benchmarks it improves pass rate by up to $2.3\%$, while in all settings maintaining the easier objectives near their already satisfied levels. Additional experiments characterize the accompanying reward tradeoffs and sensitivity to corrupted rewards.
\end{abstract}

\section{Introduction}

Reinforcement learning with verifiable rewards (RLVR) has emerged as a standard approach for improving the reasoning capabilities of large language models.
Group Relative Policy Optimization (GRPO)~\citep{shao2024deepseekmath} and its variants~\citep{guo2025deepseek} simplify this process by estimating advantages from groups of rollouts, avoiding the learned value function required by PPO~\citep{schulman2017proximal}.
At its core, however, GRPO assumes a single scalar reward for each rollout. In practice, reasoning models are often optimized for multiple objectives \citep{guo2025deepseek,fang2025thinkless}.
A response should not only be correct, but may also need to satisfy constraints on length \citep{luo2026frost,fu2025rr,feng2025efficient}, format, safety \citep{zhang2026towards,chen2025shieldagent}, or executability, etc.
The standard approach is to combine these objectives through a fixed weighted sum and then standardize the resulting scalar reward within each rollout group.
We identify two limitations of this design.
First, \emph{scalarization loses reward resolution}: different weights sum of objective rewards may produce the same scalar value and therefore receive the same advantage.
Second, \emph{fixed weights ignore objective saturation}: Each objective retains the same relative weight throughout training, regardless of how close it is to saturation. Consequently, optimization may continue to prioritize a well-optimized objective while the harder under-optimized objectives are overlooked.

To this end, we introduce \textbf{Saturation-aware Advantage Reweighting for Multi-Reward Policy Optimization} (SA-MRPO).
SA-MRPO preserves per-objective normalization while adaptively reweighting each objective according to its current degree of saturation.
Specifically, we measure saturation using the batch mean reward relative to prescribed reward bounds and reduce the relative weight of objectives whose mean rewards approach their upper bounds.
%
%
A single exponent $\gamma$ controls the strength of this saturation-aware reweighting.
We derive a threshold at which this reweighting reverses the sign of a rollout's aggregate advantage.
When saturation reweighting is disabled, SA-MRPO recovers the GDPO advantage under the same objective weights and final batch normalization \citep{liu2026gdpo}.
With one objective and a positive effective weight, its advantages are proportional to those of GRPO \citep{shao2024deepseekmath}, as detailed in Appendix~\ref{app:single_objective}.
%
%
%
%
Concurrent approaches such as DVAO~\citep{jiang2026dvao} and GD$^2$PO~\citep{liu2026gd} adapt multi-reward optimization using reward variance or advantage consistency, but do not explicitly account for how close each objective is to saturation.
Consequently, an already saturated objective can continue to exert comparable influence to one with substantially greater room for improvement. SA-MRPO addresses this limitation by making the allocation of optimization effort explicitly depend on each objective's remaining headroom, while leaving the underlying GRPO policy update unchanged.

%
%
%

To summarize, our contributions are:
\begin{itemize}[nosep,leftmargin=*]
\item We analyze how fixed aggregation of independently normalized advantages can retain substantial influence from objectives whose rewards are already near their upper bounds.

\item We propose SA-MRPO, which dynamically reweights normalized reward objectives according to their degree of optimized saturation. We establish an exact condition for reversal of a rollout's aggregate advantage and characterize the relation to GDPO and GRPO.

\item We evaluate SA-MRPO across mathematical reasoning, adaptive reasoning, and code generation, and characterize its accuracy gains, auxiliary reward tradeoffs, and sensitivity to reward corruption.


\end{itemize}


\section{Related Work}
\label{sec:related_work}
Reinforcement learning with verifiable rewards (RLVR) has become a standard paradigm for improving language-model reasoning, with GRPO~\citep{shao2024deepseekmath}, DeepSeek-R1~\citep{guo2025deepseek}, DAPO~\citep{yu2026dapo}, and REINFORCE++~\citep{hu2025reinforce++} developing increasingly effective critic-free policy optimization schemes. Extending RLVR beyond a single reward requires balancing objectives such as correctness, efficiency, helpfulness, and safety. Prior work has explored conditional preference control~\citep{zhou2024beyond}, multi-objective preference optimization~\citep{mukherjee2024multi}, Pareto-oriented optimization~\citep{li2025gradient,liu2025amopo}, and adaptive reward weighting~\citep{lu2026learning}. More recent methods study multi-reward optimization directly within group-relative policy optimization. GDPO~\citep{liu2026gdpo} independently normalizes each reward dimension before aggregation. Concurrent to our work, DVAO~\citep{jiang2026dvao} adjusts objective weights according to reward variance, and GD$^2$PO~\citep{liu2026gd} mitigates conflicts among reward-specific advantages. Other approaches consider reward correlation~\citep{ramesh2026multi}, task imbalance~\citep{liang2026don}, and alternative multi-reward optimization strategies~\citep{wang2026smopd}. The closest set of concurrent work are Dynamic Reward Weighting~\citep{lu2026learning} which adapts reward importance over training, SAW~\citep{he2026saw} which uses reward variability to estimate objective informativeness, and Focal Reward~\citep{huang2026focal} which estimates saturation for rubric-based criteria. In contrast, SA-MRPO measures saturation directly from the batch-average reward relative to its attainable range and uses it to reweight independently normalized reward advantages, allocating optimization effort toward objectives with greater remaining headroom.

\section{Preliminaries}
We study reinforcement learning for an auto-regressive language model under multiple reward objectives.
Let $\pi_\theta$ denote the policy parameterized by $\theta$.
Given a query $q$, the policy defines an auto-regressive distribution over an output sequence $o \triangleq (o_1,\ldots,o_{|o|})$ according to $\pi_\theta(o\mid q)\triangleq\prod_{t=1}^{|o|}
\pi_\theta(o_t\mid q,o_{1:t-1}),$ where $o_{1:t-1}\triangleq(o_1,\ldots,o_{t-1})$ denotes the prefix preceding token $o_t$.
At each policy update, a batch of $B$ queries
$\{q_i\}_{i=1}^{B}$ is sampled from a data distribution $\mathcal{D}$.
For each query $q_i$, a frozen behavior policy $\pi_{\theta_{\mathrm{old}}}$ generates a group of $G\geq 2$ rollouts,
{\small
\begin{align*}
o_{i,j} \sim \pi_{\theta_{\mathrm{old}}}(\cdot\mid q_i), \quad j\in\{1,\ldots,G\}.
\end{align*}}
Let $o_{i,j}\triangleq(o_{i,j,1},\ldots,o_{i,j,|o_{i,j}|})$
denote the $j$-th rollout associated with query $q_i$, and let $o_{i,j,<t}\triangleq(o_{i,j,1},\ldots,o_{i,j,t-1})$ denote its prefix before token $t$.
The policy is trained with $n$ reward objectives.
For objective $k\in\{1,\ldots,n\}$, let
$R^{(k)}$ denote its reward function and let $w_k\geq 0$ denote its prescribed weight.
The reward assigned by objective $k$ to rollout $o_{i,j}$ is $r_k^{(i,j)}\triangleq R^{(k)}(q_i,o_{i,j}).$
For a finite collection $S\triangleq \{x_1,\ldots,x_m\}$, the mean and the standard deviation of $S$ is represented as:
{\small
\begin{align*}
\mathrm{mean}(S) =\frac{1}{m}\sum_{l=1}^{m}x_l, \quad \mathrm{std}(S) = \sqrt{ \frac{1}{m}\sum_{l=1}^{m}\left(x_l-\mathrm{mean}(S)\right)^2},
\end{align*}}
respectively.
Throughout this paper, a standardized value is defined to be zero whenever its standard deviation is zero, including in the final batch normalization.

\paragraph{Group Relative Policy Optimization.}

GRPO estimates the advantage of a rollout by comparing its reward with the other rollouts generated for the same query.
With multiple reward objectives, a standard approach first combines the individual rewards into a scalar score $r_{\mathrm{sum}}^{(i,j)} \triangleq \sum_{k=1}^{n}w_k r_k^{(i,j)}.$
The resulting score is then standardized within the group generated for query $q_i$.
The GRPO advantage of rollout $o_{i,j}$ is defined as
{\small
\begin{align*}
A_{\mathrm{GRPO}}^{(i,j)} \triangleq \frac{ r_{\mathrm{sum}}^{(i,j)} - \mathrm{mean} \left\{
r_{\mathrm{sum}}^{(i,1)}, \ldots, r_{\mathrm{sum}}^{(i,G)} \right\}}{
\mathrm{std} \left\{ r_{\mathrm{sum}}^{(i,1)},
\ldots, r_{\mathrm{sum}}^{(i,G)} \right\} }.
\end{align*}}

For a clipping threshold $\epsilon>0$, define $\mathrm{clip}(x,a,c) \triangleq \max\left(a,\min(x,c)\right).$
GRPO maximizes the clipped surrogate objective
{\small
\begin{align*}
\mathcal{J}_{\mathrm{GRPO}}(\theta) = \mathbb{E}
\left[ \frac{1}{G} \sum_{j=1}^{G} \frac{1}{|o_{i,j}|} \sum_{t=1}^{|o_{i,j}|} \min \left( \rho_{i,j,t}(\theta) A_{\mathrm{GRPO}}^{(i,j)},\mathrm{clip} \left(
\rho_{i,j,t}(\theta), 1-\epsilon, 1+\epsilon
\right)
A_{\mathrm{GRPO}}^{(i,j)} \right) \right],
\end{align*}}
where the expectation is taken over
$q_i\sim\mathcal{D}$
and
$\{o_{i,j}\}_{j=1}^{G}$
sampled from
$\pi_{\theta_{\mathrm{old}}}(\cdot\mid q_i)$ and 
{\small
\begin{align*}
\rho_{i,j,t}(\theta) \triangleq \frac{
\pi_\theta \left(
o_{i,j,t} \mid q_i,o_{i,j,<t} \right)
}{
\pi_{\theta_{\mathrm{old}}} \left(
o_{i,j,t} \mid q_i,o_{i,j,<t}
\right)
}.
\end{align*}}

As in standard GRPO, a KL penalty against a fixed reference policy may additionally be included with coefficient $\beta\geq 0$.
The analysis below concerns the reward surrogate without this additional penalty.

\section{Saturation Aware Advantage Reweighting}
\label{sec:SA reward}

\begin{figure}[t]
\centering
\includegraphics[width=\linewidth]{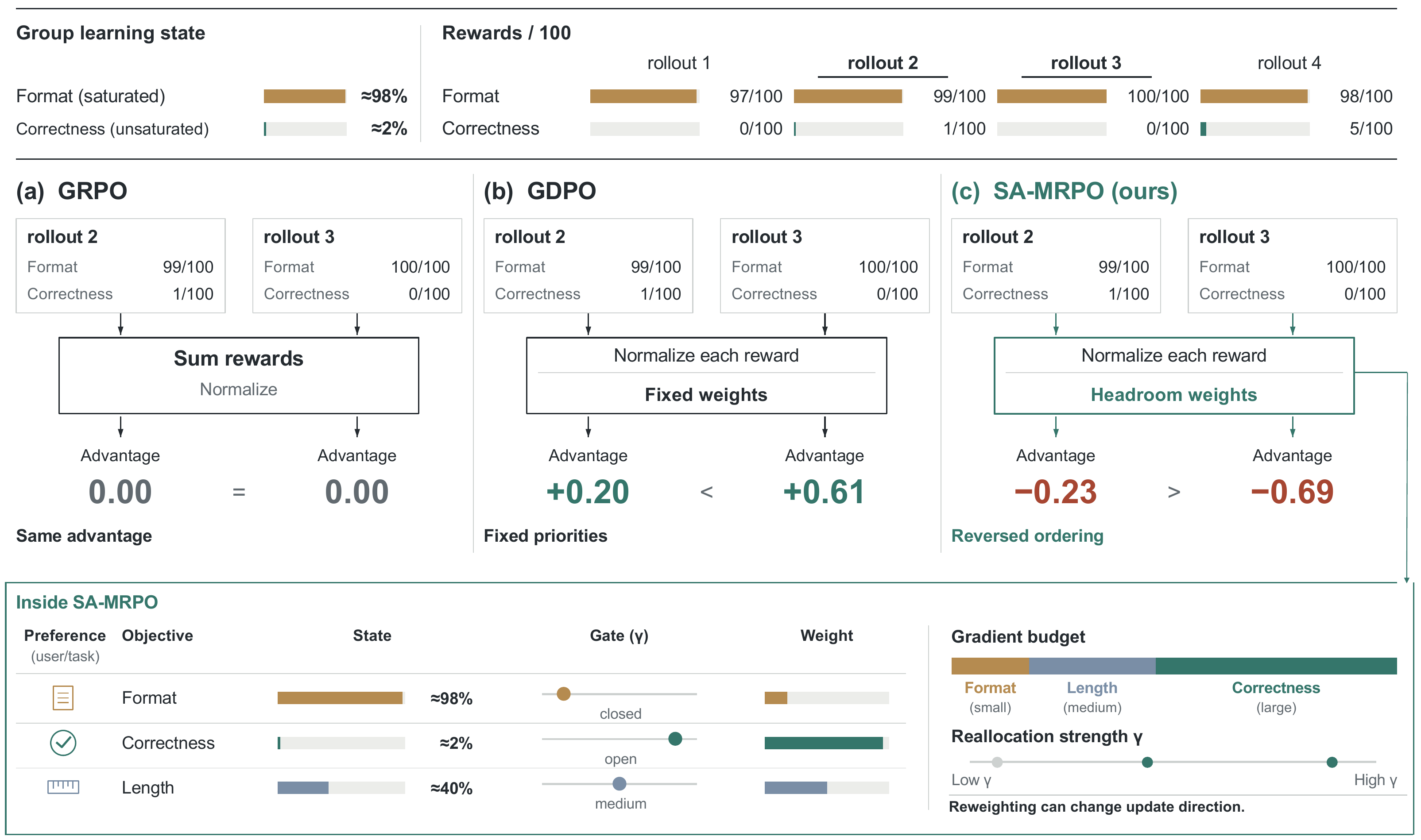}
\caption{
\textbf{Saturation-aware reweighting changes rollout preference.}
GRPO gives rollouts 2 and 3 identical advantages, while GDPO ranks rollout 3 higher despite its lower correctness reward.
SA-MRPO discounts the nearly saturated format objective, reversing both the ordering and the advantage signs relative to GDPO.
The illustrative example uses one query, four rollouts, continuous rewards in $[0,1]$ displayed out of 100, equal objective weights, and $\gamma=1$ for SA-MRPO.
}
\label{fig:saturation}
\end{figure}

The GRPO construction above combines all reward dimensions before computing the group relative advantage.
This scalarization introduces two limitations.
First, distinct reward profiles can collapse to the same scalar reward.
For example, under equal weights, $(1,0)$ and $(0,1)$ become indistinguishable.
Second, scalarization does not account for how much improvement remains for each objective.
%
%
The policy can continue favoring an already saturated objective instead of directing more of the update toward the objective that still requires improvement, as illustrated in Figure~\ref{fig:saturation}.
These two limitations motivate separating objective specific relative performance from the current optimization state of each objective.
%

\subsection{Saturation Aware Group Relative Advantage}
\label{sec:sa_advantage}

For each query $q_i$, let $r_k^{(i,j)}$ denote the reward assigned by objective $k$ to rollout $o_{i,j}$, where $i\in\{1,\ldots,B\}$, $j\in\{1,\ldots,G\}$, and $k\in\{1,\ldots,n\}$.
A multi objective policy update must account for both the relative quality of a rollout under each objective and the current optimization state of that objective.
We incorporate these two sources of information directly into a single group relative advantage.

For each objective $k$, define the group statistics associated with query $q_i$ as
{\small
\begin{align*}
\mu_k^{(i)} \triangleq \mathrm{mean} \left\{ r_k^{(i,1)},\ldots,r_k^{(i,G)} \right\},\quad
\sigma_k^{(i)} \triangleq \mathrm{std} \left\{ r_k^{(i,1)},\ldots,r_k^{(i,G)} \right\}.
\end{align*}}
To characterize the current optimization state of objective $k$, we use its average reward over the current batch, $\bar r^{(k)} \triangleq \mathrm{mean} \left\{ r_k^{(i,j)} : i\in\{1,\ldots,B\}, j\in\{1,\ldots,G\} \right\}.$
For each objective $k$, let $r_{\min}^{(k)}<r_{\max}^{(k)}$ be prescribed bounds satisfying $r_{\min}^{(k)} \le R^{(k)}(q,o) \le r_{\max}^{(k)}.$
We then define the saturation ratio of objective $k$ as
{\small
\begin{align*}
s^{(k)} \triangleq \frac{\bar r^{(k)}-r_{\min}^{(k)}}{r_{\max}^{(k)}-r_{\min}^{(k)}} \in [0,1].
\end{align*}}
A smaller $s^{(k)}$ indicates a larger gap between the observed mean reward and the prescribed upper bound.
%


Given prescribed objective weights $\{w_k\}_{k=1}^{n}$ and a saturation exponent $\gamma\geq0$, we define the saturation aware group relative advantage before final batch normalization as
{\small
\begin{align*}
\widetilde A^{(i,j)} \triangleq \sum_{k=1}^{n}\widetilde w_k A_k^{(i,j)},\quad
A_k^{(i,j)} \triangleq \frac{r_k^{(i,j)}-\mu_k^{(i)}}{\sigma_k^{(i)}},\quad
\widetilde w_k \triangleq w_k\left(1-s^{(k)}\right)^\gamma.
\end{align*}}
The term $A_k^{(i,j)}$ measures the relative quality of rollout $o_{i,j}$ under objective $k$ within the corresponding rollout group, while $\widetilde w_k$ modulates the contribution of objective $k$ according to both its prescribed importance and its current saturation level.
For $\gamma=0$, the saturation factor is defined to be one for every $s^{(k)}\in[0,1]$.
For $\gamma>0$, this factor decreases as the observed mean reward approaches the prescribed upper bound.
Thus, each objective contributes to the aggregate advantage according to its prescribed importance, its remaining room for improvement, and the relative quality of the current rollout.

Since the saturation ratios evolve during training, the scale of $\widetilde A^{(i,j)}$ may vary across policy updates.
We therefore normalize the aggregate advantages over the current batch as
{\small
\begin{align}
\widehat A_{\mathrm{SA}}^{(i,j)} \triangleq
\frac{\widetilde A^{(i,j)}-\mathrm{mean}(\mathcal A)}{\mathrm{std}(\mathcal A)},
\qquad
\mathcal A \triangleq
\left\{\widetilde A^{(i,j)}:i\in\{1,\ldots,B\},j\in\{1,\ldots,G\}\right\}.
\label{eq:batchnorm}
\end{align}}
The resulting advantage is directly used in the standard clipped group relative surrogate objective,
{\small
\begin{align*}
\mathcal{J}_{\mathrm{SA-MRPO}}(\theta) = \mathbb{E} \Bigg[
\frac{1}{G} \sum_{j=1}^{G} \frac{1}{|o_{i,j}|}
\sum_{t=1}^{|o_{i,j}|} \min \Bigg(
\rho_{i,j,t}(\theta)\widehat A_{\mathrm{SA}}^{(i,j)}, \mathrm{clip}
\left(
\rho_{i,j,t}(\theta), 1-\epsilon, 1+\epsilon
\right)
\widehat A_{\mathrm{SA}}^{(i,j)} \Bigg)
\Bigg].
\end{align*}}
Here, the expectation is over the entire batch of independently sampled queries and conditionally independent rollouts.
The query index $i$ is any fixed index in $\{1,\ldots,B\}$.
Thus, SA-MRPO changes only the construction of the rollout advantage while retaining the underlying group relative policy optimization objective.

\begin{algorithm}[t]
\caption{SA-MRPO policy update}
\label{alg:sagdpo}
\begin{algorithmic}[1]
\Require queries $\{q_i\}_{i=1}^{B}$, weights $\{w_k\}_{k=1}^{n}$, reward bounds $\{(r_{\min}^{(k)},r_{\max}^{(k)})\}_{k=1}^{n}$, saturation exponent $\gamma$, clipping threshold $\epsilon$

\State sample $\{o_{i,j}\}_{j=1}^{G} \sim \pi_{\theta_{\mathrm{old}}}(\cdot\mid q_i)$ for each $i$

\State compute $r_k^{(i,j)} = R^{(k)}(q_i,o_{i,j})$ for all $i,j,k$

\For{$k=1$ to $n$}

\State $\bar r^{(k)} \gets \mathrm{mean} \{r_k^{(i,j)}\}_{i,j}$

\State $s^{(k)} \gets \dfrac{\bar r^{(k)}-r_{\min}^{(k)}}{r_{\max}^{(k)}-r_{\min}^{(k)}}$

\EndFor

\For{each query $i$ and rollout $j$}

\State $\widetilde A^{(i,j)} \gets \displaystyle \sum_{k=1}^{n} w_k \left(1-s^{(k)}\right)^\gamma \dfrac{r_k^{(i,j)}-\mathrm{mean}\{r_k^{(i,1)},\ldots,r_k^{(i,G)}\}}{\mathrm{std}\{r_k^{(i,1)},\ldots,r_k^{(i,G)}\}}$

\EndFor

\State $\widehat A_{\mathrm{SA}}^{(i,j)} \gets \bigl(\widetilde A^{(i,j)}-\mathrm{mean}\{\widetilde A\}\bigr)/\mathrm{std}\{\widetilde A\}$ for all $i,j$

\State update $\theta$ by ascending $\nabla_\theta\mathcal{J}_{\mathrm{SA-MRPO}}(\theta)$ using $\{\widehat A_{\mathrm{SA}}^{(i,j)}\}$

\end{algorithmic}
\end{algorithm}

Algorithm~\ref{alg:sagdpo} summarizes the resulting policy update.
The saturation rule therefore provides an adaptive allocation mechanism, but changing the relative allocation across objectives also raises two questions: whether emphasizing an unsaturated objective can degrade an already optimized objective, and whether the saturation estimate itself can incorrectly represent the remaining optimization potential.

\subsection{Properties and Limitations of Saturation Aware Reweighting}
\label{sec:saturation_analysis}

The proposed construction contains static objective allocation as a special case.
When $\gamma=0$, the saturation factors are identically one and
{\small
\begin{align*}
\widetilde w_k=w_k,\quad
\widetilde A^{(i,j)}=\sum_{k=1}^{n}w_k A_k^{(i,j)},
\end{align*}}
which recovers the corresponding GDPO advantage under the same objective weights and normalization.

The saturation aware reweighting also reduces the relative contribution of the more saturated objective compared with the prescribed allocation.
Increasing $\gamma$ further shifts the relative allocation toward objectives with larger remaining reward headroom.
For any two objectives $a$ and $b$ satisfying $w_a>0$, $w_b>0$, and $s^{(a)},s^{(b)}\in[0,1)$,
{\small
\begin{align*}
\frac{\widetilde{w}_a}{\widetilde{w}_b} = \frac{w_a}{w_b} \left(
\frac{1-s^{(a)}}{1-s^{(b)}} \right)^{\gamma}.
\end{align*}}
If $s^{(a)}>s^{(b)}$, then $(1-s^{(a)})/(1-s^{(b)})<1$.
Therefore, the ratio $\widetilde{w}_a/\widetilde{w}_b$ is strictly decreasing in $\gamma$.
In particular, for every $\gamma>0$, we have $\widetilde{w}_a/\widetilde{w}_b <w_a/w_b.$
Appendix~\ref{app:single_objective} gives the relation to GRPO in the single objective setting, including groups with zero reward variance.

The reallocation is not merely a rescaling.
Figure~\ref{fig:saturation} illustrates a change in rollout preference caused by unequal saturation weights.
The following result gives an exact threshold for this effect.
\begin{proposition}[Sign reversal of a rollout advantage]
\label{prop:sign}
Fix a rollout batch with two objectives and positive weights $w_1,w_2$.
Consider a rollout satisfying $A_1^{(i,j)}>0>A_2^{(i,j)}$.
Let $b_k\triangleq1-s^{(k)}$ for $k\in\{1,2\}$, and suppose $0<b_1<b_2\le1$.
Define $a\triangleq w_1A_1^{(i,j)}$, $
c\triangleq w_2|A_2^{(i,j)}|$, $\chi\triangleq a/c>1.$
Then $\widetilde A^{(i,j)}(\gamma)=ab_1^\gamma-cb_2^\gamma$ changes sign exactly once, at
{\small
\begin{align}
\label{eq:gamma_star}
    \gamma^\star = \frac{\ln\chi}{\ln(b_2/b_1)}.
\end{align}}
It is positive for $0\le\gamma<\gamma^\star$, zero at $\gamma^\star$, and negative for $\gamma>\gamma^\star$.
\end{proposition}

Appendix~\ref{app:normalization} proves Proposition~\ref{prop:sign} and establishes that final standardization preserves the resulting advantage sign.
This result concerns a rollout advantage and does not imply that the full parameter update reverses direction.
For a fixed $\chi$, Equation~\ref{eq:gamma_star} shows that the reversal threshold decreases as $\ln(b_2/b_1)$ increases.

Saturation aware reweighting reallocates optimization emphasis according to remaining reward headroom, but it does not impose a constraint that previously optimized objectives must be preserved.
Conflicting objective gradients can produce a descent direction for an objective surrogate, as detailed in Appendix~\ref{app:objective_conflict}.
This analysis concerns the initial surrogate direction for a fixed rollout batch and does not establish a decrease in expected reward.
A separate consideration is that nominal reward headroom does not necessarily coincide with optimizable headroom.
By construction, $1-s^{(k)}$ measures the fraction of the prescribed reward range that remains unrealized.
Appendix~\ref{app:corrupted_rewards} examines sensitivity to corrupted length rewards.

\section{Experiments}
\label{sec:experiment}
We evaluate SA-MRPO with three questions in mind: whether saturation aware reweighting improves policy optimization under multiple reward objectives, whether the same behavior extends beyond mathematical reasoning, and whether the observed objective weights and reward tradeoffs agree with the intended saturation aware allocation.


\subsection{Experimental Setup}
\label{sec:exp_setup}
\paragraph{Training protocol.}
Unless otherwise specified, all experiments are implemented with \texttt{verl} and use vLLM for rollout generation.
The DeepScaleR experiments use $G=4$ responses per prompt, one training epoch, a global batch size of $256$, and a maximum response length of $4096$ tokens.
The additional baseline comparison uses a mini batch size of $64$ and a learning rate of $10^{-6}$.
Appendix~\ref{app:experimental_protocols} specifies the coding configuration.
Mathematical and adaptive reasoning experiments are conducted on DeepScaleR-Preview \cite{luo2025deepscaler}.
The model architecture, reward construction, and task-specific deviations from this protocol are described in the corresponding subsections.

\paragraph{Evaluation protocol.}
Mathematical evaluation uses vLLM with a maximum generation length
of $4096$ tokens.
Tables~\ref{tab:qwen_reward_settings}, \ref{tab:sagdpo_gdpo_adaptive_reasoning}, and~\ref{tab:sagdpo_gamma_sweep_3b} use $16$ sampled responses per problem.
Accuracy is the mean correctness of these responses.
Each configuration uses one training run.
Table~\ref{tab:controlled_adaptive} uses greedy decoding and multiple training seeds.
Code evaluation uses $16$ sampled responses per problem with a maximum generation length of $2048$ tokens.
Task specific auxiliary metrics are introduced together with the corresponding experiments.
In all evaluation tables, ``Base'' denotes the corresponding model before RL training.

\begin{figure}
    \centering
    \includegraphics[width=\linewidth]{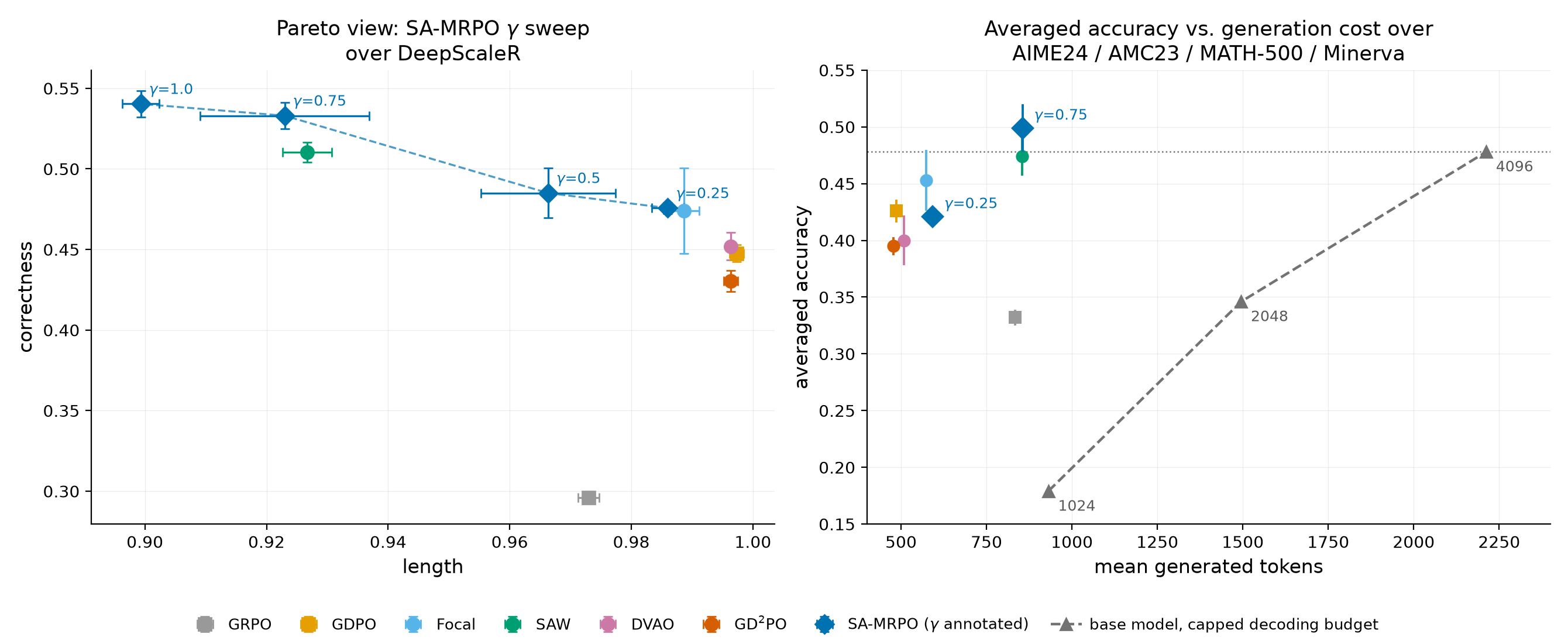}
    \caption{Reward tradeoffs and generation cost in controlled adaptive reasoning. \emph{Left:} mean evaluation rewards over three seeds, with sample standard deviations. SA-MRPO settings are connected in $\gamma$ order. \emph{Right:} mean benchmark accuracy versus generated tokens. Trained methods use a 4096 token cap and two or three seeds. The base model is evaluated with caps of 1024, 2048, and 4096 tokens.
    Vertical error bars show sample standard deviations across training seeds.}
    \label{fig:pareto}
\end{figure}

\subsection{Mathematical Reasoning}
\label{sec:math_reasoning}

We first investigate whether saturation-aware reweighting improves general
multi-objective reasoning performance. We train Qwen2.5-3B-Instruct and
Qwen2.5-7B-Instruct following the training protocol described in
Section~\ref{sec:exp_setup}.
We consider both two and three objective settings.
The two objective setting optimizes correctness and response length compliance, while the three objective setting additionally includes a format objective.
%
%
The length objective favors shorter responses, while the format objective checks the prescribed \texttt{<think>} and \texttt{<answer>} tag structure.
Appendix~\ref{app:reward_definitions} defines the correctness, length, and format rewards.
%
%
For Table~\ref{tab:qwen_reward_settings}, the quadratic length reward uses $L_{\min}=0$ and $L_{\max}=4000$.
The 3B configuration with two objectives uses $\gamma=0.25$.

We evaluate on AIME24~\citep{aime24},
AMC23~\footnote{https://huggingface.co/datasets/math-ai/amc23},
MATH500~\citep{hendrycks2021math},
Minerva Math~\footnote{https://huggingface.co/datasets/math-ai/minervamath},
and OlympiadBench~\citep{he2024olympiadbench}.
In addition to accuracy, we report \textsc{Exceed}, defined as the fraction of generated responses whose length exceeds the 4000 token budget.
Table~\ref{tab:qwen_reward_settings} compares SA-MRPO with GDPO for three configurations: 7B with three objectives, and 3B with two or three objectives.
SA-MRPO achieves higher accuracy on four of the five benchmarks in each configuration, giving $12$ improvements across the $15$ comparisons.
The exceptions are AMC23 for the 7B model and OlympiadBench for both 3B configurations.
The reported exceedance rates for SA-MRPO are at most $1.7\%$, although they are higher than GDPO's in $10$ comparisons.
These results indicate an accuracy gain accompanied by a tradeoff in length budget compliance.

\begin{table}[t]
\centering
\scriptsize
\setlength{\tabcolsep}{3.5pt}
\renewcommand{\arraystretch}{1.12}
\setlength{\aboverulesep}{0pt}
\setlength{\belowrulesep}{0pt}
\setlength{\extrarowheight}{0.75ex}
\caption{
Comparison of SA-MRPO, GDPO, and the base model on mathematical reasoning with two and three reward objectives. Higher accuracy is better, while \textsc{Exceed} measures the fraction of responses exceeding the 4,000-token budget and is lower is better. The same Qwen2.5-3B-Instruct base model is used for both reward settings.
}
\resizebox{\textwidth}{!}{
\begin{tabular}{llccc|ccccc}
\toprule
\multirow{3}{*}{Benchmark}
& \multirow{3}{*}{Metric}
& \multicolumn{3}{c|}{Qwen2.5-7B-Instruct}
& \multicolumn{5}{c}{Qwen2.5-3B-Instruct} \\
\cmidrule(lr){3-5}
\cmidrule(lr){6-10}
&
& \multicolumn{3}{c|}{Three objectives}
& \multirow{2}{*}{Base}
& \multicolumn{2}{c}{$\mathcal{R}_{\mathrm{correct}}+\mathcal{R}_{\mathrm{length}}$}
& \multicolumn{2}{c}{$\mathcal{R}_{\mathrm{correct}}+\mathcal{R}_{\mathrm{length}}+\mathcal{R}_{\mathrm{format}}$} \\
\cmidrule(lr){3-5}
\cmidrule(lr){7-8}
\cmidrule(lr){9-10}
&
& Base & GDPO & SA-MRPO
&
& GDPO$_{\mathrm{2obj}}$ & SA-MRPO$_{\mathrm{2obj}}$
& GDPO$_{\mathrm{3obj}}$ & SA-MRPO$_{\mathrm{3obj}}$ \\
\midrule

\multirow{2}{*}{AIME24}
& Acc $\uparrow$
& 11.7\% & 11.5\% & \textbf{16.5\%}
& 0.6\%
& 5.0\% & \textbf{8.5\%}
& 6.7\% & \textbf{8.1\%}\\
& Exceed $\downarrow$
& 3.5\% & \textbf{1.2\%} & 1.5\%
& 6.2\%
& \textbf{0.0\%} & 0.6\%
& \textbf{0.6\%} & 1.7\% \\
\midrule

\multirow{2}{*}{Minerva}
& Acc $\uparrow$
& 16.1\% & 24.2\% & \textbf{24.8\%}
& 6.7\%
& 16.2\% & \textbf{16.6\%}
& 16.9\% & \textbf{18.1\%} \\
& Exceed $\downarrow$
& 0.4\% & 0.1\% & \textbf{0.0\%}
& 0.4\%
& \textbf{0.1\%} & \textbf{0.1\%}
& \textbf{0.1\%} & \textbf{0.1\%} \\
\midrule

\multirow{2}{*}{AMC23}
& Acc $\uparrow$
& 41.1\% & \textbf{44.6\%} & 43.5\%
& 10.7\%
& 33.2\% & \textbf{34.9\%}
& 31.5\% & \textbf{35.2\%}\\
& Exceed $\downarrow$
& \textbf{1.0\%} & 1.2\% & 1.1\%
& 1.7\%
& \textbf{0.0\%} & 0.1\%
& \textbf{0.4\%} & 0.6\% \\
\midrule

\multirow{2}{*}{MATH500}
& Acc $\uparrow$
& 50.0\% & 64.2\% & \textbf{67.7\%}
& 26.3\%
& 57.1\% & \textbf{58.2\%}
& 58.9\% & \textbf{59.5\%} \\
& Exceed $\downarrow$
& 0.7\% & \textbf{0.0\%} & 0.1\%
& 0.6\%
& \textbf{0.0\%} & 0.1\%
& \textbf{0.1\%} & \textbf{0.1\%} \\
\midrule

\multirow{2}{*}{Olympiad}
& Acc $\uparrow$
& 23.8\% & 25.3\% & \textbf{26.1\%}
& 4.5\%
& \textbf{20.6\%} & 19.3\%
& \textbf{20.6\%} & 20.0\% \\
& Exceed $\downarrow$
& 2.5\% & \textbf{0.2\%} & 0.7\%
& 3.3\%
& \textbf{0.1\%} & 0.3\%
& \textbf{0.2\%} & 0.6\% \\
\bottomrule
\end{tabular}
}

\label{tab:qwen_reward_settings}
\end{table}

\subsection{Adaptive Reasoning}
\label{sec:adaptive_reasoning}

The previous experiment evaluates SA-MRPO under standard multi-objective reward constructions.
We train DeepSeek-R1-Distill-Qwen-7B with the same correctness and response length compliance objectives used in the two objective setting above.
No learned judge or external reward model is used.
%
%
The correctness reward is one when the parsed answer matches the reference answer and zero otherwise.
With $L_{\min}=0$, the length reward decreases quadratically with response length until it reaches zero at $L_{\max}$, and remains zero beyond that threshold.
Appendix~\ref{app:reward_definitions} gives both reward definitions.
We evaluate on the same five mathematical reasoning benchmarks as in Section~\ref{sec:math_reasoning}.
We report accuracy together with \textsc{Len}, the average number of generated tokens.

\begin{table}[t]
\centering

\begin{minipage}[t]{0.49\linewidth}
\centering
\scriptsize
\setlength{\tabcolsep}{3pt}
\renewcommand{\arraystretch}{1.10}

\caption{Performance comparison on adaptive reasoning, reporting both accuracy and mean response length.}
\label{tab:sagdpo_gdpo_adaptive_reasoning}

\resizebox{\linewidth}{!}{
\begin{tabular}{lccccc}
\toprule
\multirow{2}{*}{Benchmark}
& \multicolumn{2}{c}{SA-MRPO}
& \multicolumn{2}{c}{GDPO}
& \multirow{2}{*}{$\Delta$ Acc.} \\
\cmidrule(lr){2-3}
\cmidrule(lr){4-5}
& Acc. $\uparrow$ & Len. $\downarrow$ & Acc. $\uparrow$ & Len. $\downarrow$ & \\
\midrule

AIME24
& \textbf{7.3} & 804
& 5.2 & 566
& +2.1 \\

Minerva
& \textbf{15.9} & 277
& 15.4 & 214
& +0.5 \\

AMC23
& \textbf{37.5} & 417
& 28.3 & 290
& +9.2 \\

MATH500
& \textbf{51.5} & 270
& 47.1 & 187
& +4.4 \\

Olympiad
& \textbf{20.9} & 529
& 18.1 & 406
& +2.8 \\

\midrule
Average
& \textbf{26.6} & 459
& 22.8 & 333
& +3.8 \\

\bottomrule
\end{tabular}
}
\end{minipage}
\hfill
\begin{minipage}[t]{0.49\linewidth}
\centering
\scriptsize
\setlength{\tabcolsep}{3pt}
\renewcommand{\arraystretch}{1.10}

\captionof{table}{Performance comparison on Code generation and reasoning on Qwen2.5-7B-Instruct.}
\label{tab:qwen-code}

\resizebox{\linewidth}{!}{
\begin{tabular}{llccc}
\toprule
& & Base & GDPO & SA-MRPO \\
\midrule

\multirow{2}{*}{APPS}
& Pass $\uparrow$
& 43.8\% & 53.2\% & \textbf{53.8\%} \\
& Bug $\downarrow$
& 19.6\% & \textbf{8.5\%} & 9.9\% \\

\midrule
\multirow{2}{*}{CodeCont.}
& Pass $\uparrow$
& 12.4\% & 19.2\% & \textbf{20.6\%} \\
& Bug $\downarrow$
& 32.9\% & \textbf{15.5\%} & \textbf{15.5\%} \\

\midrule
\multirow{2}{*}{Codeforces}
& Pass $\uparrow$
& 8.7\% & 10.6\% & \textbf{12.9\%} \\
& Bug $\downarrow$
& 34.4\% & \textbf{8.6\%} & 9.0\% \\

\midrule
\multirow{2}{*}{TACO}
& Pass $\uparrow$
& 29.0\% & \textbf{36.0\%} & 35.6\% \\
& Bug $\downarrow$
& 23.1\% & \textbf{11.0\%} & 12.4\% \\

\bottomrule
\end{tabular}
}
\end{minipage}

\end{table}

Table~\ref{tab:sagdpo_gdpo_adaptive_reasoning} shows that SA-MRPO achieves higher accuracy than GDPO on all five benchmarks, raising the average from $22.8\%$ to $26.6\%$.
The largest improvement occurs on AMC23, where accuracy increases from $28.3\%$ to $37.5\%$.
At the same time, SA-MRPO produces moderately longer responses, increasing the average response length from $333$ to $459$ tokens.

We compare SA-MRPO with GRPO, GDPO, and adaptive reward allocation baselines.
The controlled length rewards use
$(L_{\min},L_{\max})=(1024,2048)$ for DeepSeek-R1-Distill-Qwen-7B.
These experiments use greedy decoding and are evaluated separately from the sampled results in Tables~\ref{tab:qwen_reward_settings} and~\ref{tab:sagdpo_gdpo_adaptive_reasoning}.
%
%
Figure~\ref{fig:reward_trajectories} shows how the tradeoff between correctness and length reward develops during training.
Along the plotted GDPO trajectory, substantial correctness gains appear only near the upper bound of the length reward.
SA-MRPO with $\gamma=0.75$ and SAW turn toward higher correctness while the length reward remains farther below that bound.
The trajectories therefore show different patterns of balancing correctness and length during training.

\begin{table}[t]
\centering
\small
\caption{Adaptive reasoning with greedy decoding and a maximum generation length of $4096$ tokens.
Accuracy is averaged equally over AIME24, AMC23, MATH500, and Minerva.
Uncertainty denotes the sample standard deviation across training seeds; Appendix~\ref{app:experimental_protocols} lists the number of seeds per configuration.
The Focal baseline is adapted to pointwise rewards.
Bold values identify the highest means among trained configurations.
%
}
\label{tab:controlled_adaptive}
\setlength{\tabcolsep}{3pt}
\begin{tabular}{lcccccc}
\toprule
Method & AIME24 & AMC23 & MATH500 & Minerva & Average & Len. $\downarrow$ \\
\midrule
Base & 50.0 & 62.6 & 60.0 & 18.4 & 47.8 & 2212 \\
GRPO & 8.3 & 45.8 & 59.6 & 18.9 & $33.2\pm0.7$ & 834 \\
GDPO & 22.2 & 51.4 & 70.9 & 25.7 & $42.6\pm1.0$ & 487 \\
Focal adaptation & 25.6 & 57.0 & 71.9 & 26.7 & $45.3\pm2.7$ & 574 \\
SAW & 25.6 & 61.9 & 75.1 & 27.2 & $47.4\pm1.7$ & 855 \\
DVAO & 11.1 & 51.0 & 71.5 & 26.6 & $40.0\pm2.2$ & 508 \\
GD$^2$PO & 15.0 & 50.0 & 68.7 & 24.4 & $39.5\pm0.8$ & 478 \\
SA-MRPO, $\gamma=0.75$ & \textbf{28.3} & \textbf{65.1} & \textbf{77.0} & \textbf{29.2} & $\mathbf{49.9}\pm2.1$ & 856 \\
\bottomrule
\end{tabular}
\end{table}

In Table~\ref{tab:controlled_adaptive}, SA-MRPO has the highest mean accuracy on each of the four benchmarks among the trained configurations.
Compared with SAW, its average accuracy increases from $47.4\%$ to $49.9\%$ at almost the same mean generation length, $856$ versus $855$ tokens.
Figure~\ref{fig:pareto} separates the reward tradeoff from the generation cost.
In the left panel, increasing $\gamma$ moves SA-MRPO toward higher correctness while accepting a lower length reward.
In the right panel, SA-MRPO with $\gamma=0.75$ achieves higher benchmark accuracy than SAW at nearly identical mean generation length.
The contrast between $\gamma=0.25$ and $\gamma=0.75$ motivates the examination of saturation strength in Section~\ref{sec:gamma_ablation}.

%
%
%

\subsection{Code Generation}
\label{sec:coding_reasoning}

We next evaluate whether saturation aware reweighting extends beyond mathematical reasoning.
We train Qwen2.5-7B-Instruct on the Eurus-2-RL dataset \cite{cui2025process} using two rule based reward objectives: test case pass rate and training runs for three epochs.
GDPO and SA-MRPO use identical training data, reward functions, and optimization hyperparameters and differ only in the construction of the aggregate advantage.
The test case pass rate reward is the fraction of tests passed by a generated program.
The executability reward is one if the program compiles and runs without errors on all test cases, and zero otherwise.
A program can therefore receive full executability reward while still producing incorrect outputs.
Appendix~\ref{app:reward_definitions} provides the reward formulas and evaluation settings.
Executability captures a basic validity requirement that can become satisfied before full functional correctness.
The pass rate objective is more demanding because it requires the generated program to produce correct behavior across the test cases.
This setting therefore provides a qualitatively different instance in which one objective may become substantially easier to satisfy than another.

We evaluate on APPS \cite{hendrycks2021measuring}, CodeContests \cite{li2022competition}, Codeforces, and TACO \cite{li2023taco}.
We report \textsc{Pass}, the average test case pass rate, and \textsc{Bug}, the fraction of generated programs that encounter compilation or runtime errors.
Table~\ref{tab:qwen-code} shows that both methods substantially improve functional correctness and executability over the base model.
On CodeContests, SA-MRPO raises the pass rate from $19.2\%$ to $20.6\%$, while both methods have a reported bug rate of $15.5\%$.
On APPS and Codeforces, higher pass rates accompany higher bug rates; on TACO, SA-MRPO has a lower pass rate and a higher bug rate than GDPO.
%
%
These results demonstrate improvements in test case performance beyond mathematical reasoning on three benchmarks, with the accompanying effect on executability depending on the benchmark.

\subsection{Effect of the Saturation Strength}
\label{sec:gamma_ablation}

We finally examine whether the saturation exponent $\gamma$ controls the allocation of optimization effort in the manner predicted by the SA-MRPO weighting rule.
Recall that $\widetilde{w}_k = w_k \left( 1-s^{(k)} \right)^\gamma,$ so increasing $\gamma$ more aggressively suppresses objectives with large saturation
$s^{(k)}$.
Figure~\ref{fig:pareto} reports the adaptive reward tradeoff across saturation exponents.
Increasing $\gamma$ from $0.75$ to $1$ changes mean correctness from $0.533$ to $0.540$ and mean length reward from $0.923$ to $0.899$.
We use $\gamma=0.75$ as an empirical compromise in this setting.
Appendix~\ref{app:gamma_sweeps} reports the separate sampled 3B study, where $\gamma=0.5$ has the highest average benchmark accuracy among the evaluated settings.

\section{Conclusion}

We studied multi reward policy optimization from the perspective of how optimization effort is allocated across objectives at different stages of training.
While reward decoupling preserves information from individual reward dimensions, it does not distinguish between objectives that remain difficult and objectives that are already close to saturation.
We introduced SA-MRPO, which addresses this limitation by adapting each objective's contribution according to its current saturation while retaining the standard GRPO policy update.
Across mathematical reasoning, adaptive reasoning, and coding tasks, SA-MRPO improves the primary objective in many evaluated configurations, with accompanying tradeoffs in auxiliary rewards.
The corruption experiments also show sensitivity to reward noise.
The ablation over the saturation exponent further shows that this reallocation is controllable, exposing a direct tradeoff between improving under optimized objectives and preserving saturated auxiliary objectives.
These results suggest that effective multi reward policy optimization should account not only for the relative scale of reward signals, but also for how much useful improvement remains in each objective.

\newpage
\subsection*{AI use statement}
In this work, we have not used generative AI tools for any tasks with required disclosure.
We used generative AI tools solely for language editing to improve the grammar, clarity, and readability of the manuscript.
We have reviewed all work assisted by generative AI.
Specifically, we checked all suggested language edits to ensure that they preserved the intended meaning and technical accuracy.
We take responsibility for the final content of this work, including text, claims or artifacts produced with the aid of generative AI.

\subsection*{Ethics statement}

This work studies multi-reward reinforcement learning using existing open-source language models, training datasets, and publicly available mathematical reasoning and code-generation benchmarks. We introduce no new human-subject data collection, crowdsourcing, or personally identifiable information. The proposed SA-MRPO method modifies only the construction of rollout advantages by adaptively reweighting existing reward objectives according to their observed saturation, while retaining the underlying group-relative policy optimization procedure. The experiments rely primarily on verifiable or rule-based rewards, including answer correctness, response length, output format, test-case pass rate, and code executability, rather than newly collected human preference annotations. As with other post-training methods for language models, improved reasoning and code-generation capability may inherit or amplify limitations of the underlying models, datasets, reward functions, and evaluation benchmarks. We are not aware of additional ethical concerns specific to SA-MRPO beyond these general considerations.

\subsection*{Reproducibility statement}

We provide the model, training, reward, and evaluation configurations required to reproduce our experiments.
Section~\ref{sec:exp_setup} specifies the main training and evaluation protocol, including the base models, rollout count, batch size, training epochs, response-length limits, and decoding settings. The construction of SA-MRPO, including per-objective normalization, saturation estimation, adaptive weights, final batch normalization, and the saturation exponent $\gamma$, is fully specified in Appendix~\ref{app:gamma_sweeps} and Proposition~\ref{prop:sign}.
Appendix~\ref{app:experimental_protocols}
provides the detailed training and evaluation configurations, reward definitions, reward bounds, length thresholds, and implementations of the comparison baselines.
Additional ablations report the effects of the saturation exponent and corrupted reward signals. We will release the training and evaluation code, configuration files, and model checkpoints used in the reported experiments to facilitate reproduction of the results.

\bibliography{ref}
\bibliographystyle{plainnat}

\appendix
\section{Additional Mathematical Analysis}
\label{app:math}

This appendix establishes sign preservation under final normalization, derives the relation to GDPO and GRPO, and characterizes the effect of conflicting objective gradients on the initial surrogate direction.

\subsection{Centering and Proof of Proposition~\ref{prop:sign}}
\label{app:normalization}

Each objective advantage has zero mean within its rollout group.
When the group reward variance is zero, this statement follows from the zero variance convention.
Since the objective weights are shared across the batch,
\[
\mathrm{mean}(\mathcal A)
=
\frac1{BG}\sum_{i=1}^{B}\sum_{j=1}^{G}
\sum_{k=1}^{n}\widetilde w_k A_k^{(i,j)}
=
\sum_{k=1}^{n}\widetilde w_k
\frac1B\sum_{i=1}^{B}
\frac1G\sum_{j=1}^{G}A_k^{(i,j)}
=
0.
\]
If any aggregate advantage is nonzero, the batch standard deviation is positive.
Final standardization then divides every aggregate advantage by the same positive number and preserves its sign.
If the batch standard deviation is zero, all aggregate and final advantages are zero.

Under the assumptions of Proposition~\ref{prop:sign},
\[
\widetilde A^{(i,j)}(\gamma)
=
ab_1^\gamma-cb_2^\gamma
=
cb_2^\gamma
\left[
\chi\left(\frac{b_1}{b_2}\right)^\gamma-1
\right].
\]
The prefactor is positive for every $\gamma\ge0$.
Since $0<b_1/b_2<1$, the bracketed expression is continuous and strictly decreasing.
It equals $\chi-1>0$ at $\gamma=0$ and tends to $-1$ as $\gamma$ tends to infinity.
Its unique zero satisfies
\[
\chi\left(\frac{b_1}{b_2}\right)^\gamma=1
\quad\Longleftrightarrow\quad
\gamma
=
\frac{\ln\chi}{\ln(b_2/b_1)}
=
\gamma^\star.
\]
Strict monotonicity gives the three sign cases in the proposition.
The centering argument above shows that final standardization preserves these cases.

\subsection{Relation to GDPO and the Single Objective Case}
\label{app:single_objective}

When $\gamma=0$, every saturation factor is defined to be one.
The aggregate advantage is therefore
\[
\widetilde A^{(i,j)}
=
\sum_{k=1}^{n}w_k A_k^{(i,j)}.
\]
With the same prescribed weights and final normalization, this is the GDPO construction.

Now suppose that there is one objective and its effective weight satisfies $\widetilde w_1>0$.
Positive scaling of the reward does not change its group standardization, so
\[
A_{\mathrm{GRPO}}^{(i,j)}=A_1^{(i,j)}.
\]
Let $M\in\{0,\ldots,B\}$ be the number of rollout groups with positive reward variance.
Population standardization gives
\[
\frac1G\sum_{j=1}^{G}\left(A_1^{(i,j)}\right)^2
=
\begin{cases}
1,&\sigma_1^{(i)}>0,\\
0,&\sigma_1^{(i)}=0.
\end{cases}
\]
Each group has zero mean, and hence
\[
\frac1{BG}\sum_{i=1}^{B}\sum_{j=1}^{G}
\left(A_1^{(i,j)}\right)^2
=
\frac MB.
\]
Consequently,
\[
\mathrm{std}(\mathcal A)
=
\widetilde w_1\sqrt{\frac MB}.
\]
For $M>0$,
\[
\widehat A_{\mathrm{SA}}^{(i,j)}
=
\frac{\widetilde w_1A_1^{(i,j)}}{\widetilde w_1\sqrt{M/B}}
=
\sqrt{\frac BM}\,A_{\mathrm{GRPO}}^{(i,j)}.
\]
The advantages coincide when $M=B$.
When $M=0$, both methods assign zero advantages.
These identities use population standard deviations and the stated zero variance convention.
For $0<M<B$, final batch normalization introduces the factor
$\sqrt{B/M}$, so the single-objective advantages are proportional to those of GRPO.

\subsection{Initial Surrogate Direction under Objective Conflict}
\label{app:objective_conflict}

Fix the rollout batch, rewards, advantages, and saturation weights.
Assume that every sampled token probability is continuously differentiable near $\theta_{\mathrm{old}}$.
Define the empirical surrogate $J_k(\theta)$ as
\[
J_k(\theta)
\triangleq
\frac1{BG}
\sum_{i=1}^{B}\sum_{j=1}^{G}
\frac{A_k^{(i,j)}}{|o_{i,j}|}
\sum_{t=1}^{|o_{i,j}|}\rho_{i,j,t}(\theta),
\qquad
g_k\triangleq\nabla_\theta J_k(\theta_{\mathrm{old}}).
\]
Let
\[
d\triangleq\sum_{k=1}^{n}\widetilde w_k g_k,
\qquad
\sigma_A\triangleq\mathrm{std}(\mathcal A).
\]
First suppose that $\sigma_A>0$.
Let $\widehat{\mathcal J}_{\mathrm{SA}}(\theta)$ denote the empirical
reward surrogate from Section~\ref{sec:sa_advantage},
averaged over the fixed batch.
Since every sampled likelihood ratio equals one at $\theta_{\mathrm{old}}$, continuity and the finiteness of the batch imply that all ratios remain inside the clipping interval in a common neighborhood of $\theta_{\mathrm{old}}$.
Using Appendix~\ref{app:normalization}, the surrogate in this neighborhood satisfies
\[
\widehat{\mathcal J}_{\mathrm{SA}}(\theta)
=
\frac1{\sigma_A}
\sum_{k=1}^{n}\widetilde w_k J_k(\theta).
\]
Therefore,
\[
\nabla_\theta
\widehat{\mathcal J}_{\mathrm{SA}}(\theta_{\mathrm{old}})
=
\frac1{\sigma_A}
\sum_{k=1}^{n}\widetilde w_k g_k
=
\frac d{\sigma_A}.
\]

For objective $a$, its directional derivative along $d$ is
\[
DJ_a(\theta_{\mathrm{old}})[d]
\triangleq
g_a^\top d
=
\widetilde w_a\|g_a\|^2
+
\sum_{k\ne a}\widetilde w_k g_a^\top g_k,
\]
where $\|\cdot\|$ is the Euclidean norm.
This derivative is negative precisely when
\begin{equation}
\sum_{k\ne a}\widetilde w_k g_a^\top g_k
<
-\widetilde w_a\|g_a\|^2.
\label{eq:objective_conflict}
\end{equation}
Because $\sigma_A>0$, the derivative along the empirical surrogate gradient has the same sign.

If $\sigma_A=0$, all aggregate advantages are zero by the centering identity.
The final advantages and empirical reward surrogate gradient are then zero.

This analysis concerns the initial reward surrogate direction for a fixed batch without an additional reference policy penalty.
It does not establish a decrease in expected reward or characterize arbitrary optimizer transformations and later updates after clipping becomes active.

\section{Experimental Protocols and Implementations}
\label{app:experimental_protocols}

\subsection{Training and Evaluation}

The mathematical experiments use DeepScaleR-Preview with \texttt{verl} and vLLM.
DeepScaleR training uses $G=4$ responses per prompt, one epoch, a global batch size of $256$, and a maximum response length of $4096$ tokens.
The adaptive baseline comparison in Table~\ref{tab:controlled_adaptive} uses a mini batch size of $64$, a learning rate of $10^{-6}$, a PPO clipping threshold of $0.2$, and a gradient norm limit of $1.0$.
The reported hardware is four A100 GPUs with $80$ GB memory each.
The additional adaptive GRPO configuration has a format reward and a KL term incorporated into its reward, in addition to correctness and length.
This difference prevents the GRPO comparisons from isolating advantage construction under identical objectives.
The additional mathematical configurations of GDPO, Focal, SAW, DVAO, GD$^2$PO, and SA-MRPO use no KL penalty.

The sampled mathematical evaluations in Tables~\ref{tab:qwen_reward_settings}, \ref{tab:sagdpo_gdpo_adaptive_reasoning}, and~\ref{tab:sagdpo_gamma_sweep_3b} use $16$ responses per problem, a nucleus sampling threshold of $0.95$, and a generation cap of $4096$ tokens.
Each configuration corresponds to one training run.
Accuracy is the mean correctness of the sampled responses, not the probability that at least one response is correct.
The five benchmarks are AIME24, AMC23\footnote{\url{https://huggingface.co/datasets/math-ai/amc23}}, MATH500, Minerva\footnote{\url{https://huggingface.co/datasets/math-ai/minervamath}}, and OlympiadBench.
The additional comparisons use greedy decoding on the first four benchmarks and omit OlympiadBench.
For each seed, average accuracy gives equal weight to the four benchmark accuracies.
Reported sample standard deviations are computed across these seed averages.
The DeepScaleR evaluation reward and the external benchmark average are distinct metrics.

\subsection{Reward Definitions and Auxiliary Metrics}
\label{app:reward_definitions}

All reward objectives considered by SA-MRPO have prescribed bounds zero and one.
Let $y(q)$ denote the reference answer for query $q$, and let $\mathrm{Extract}(o)$ denote the answer parsed from response $o$.
The correctness reward is defined by
\[
\mathcal R_{\mathrm{correct}}(q,o)
\triangleq
\mathbf 1\{\mathrm{Extract}(o)=y(q)\},
\]
where $\mathbf 1\{E\}$ equals one when condition $E$ holds and zero otherwise.
For thresholds $0\le L_{\min}<L_{\max}$, the length reward is
\[
\mathcal R_{\mathrm{length}}(o)
\triangleq
1-\left[
\mathrm{clip}\left(
\frac{|o|-L_{\min}}{L_{\max}-L_{\min}},0,1
\right)
\right]^2,
\]
where $\mathrm{clip}(x,0,1)\triangleq\max\{0,\min\{x,1\}\}$.
The reward equals one for $|o|\le L_{\min}$ and zero for $|o|\ge L_{\max}$.
When $L_{\min}=0$, there is no positive length interval with maximal reward.
The metric \textsc{Exceed} is the fraction of generated responses with more than $4000$ tokens and is distinct from the mean length reward.

Table~\ref{tab:length_configurations} lists length thresholds for the mathematical comparisons.
The sampled adaptive experiment in Table~\ref{tab:sagdpo_gdpo_adaptive_reasoning} uses $L_{\min}=0$.
The 3B configuration with two objectives in Table~\ref{tab:qwen_reward_settings} uses $\gamma=0.25$.

\begin{table}[t]
\centering
\small
\caption{Length reward thresholds for the mathematical comparisons.}
\label{tab:length_configurations}
\begin{tabular}{lcc}
\toprule
Configuration & $L_{\min}$ & $L_{\max}$ \\
\midrule
Sampled mathematical reasoning & $0$ & $4000$ \\
Additional adaptive comparison & $1024$ & $2048$ \\
Adaptive reward sweep & $1024$ & $2048$ \\
Corrupted reward experiment & $1000$ & $4000$ \\
\bottomrule
\end{tabular}
\end{table}

For a tag string $\tau$, let $N_\tau(o)$ denote the number of occurrences of that string in response $o$.
The format reward $\mathcal R_{\mathrm{format}}(o)$ is defined as
    \[
    \mathcal{R}_{\mathrm{format}}(o)
    =
    \begin{cases}
        1,
        &
        \begin{aligned}
        &\text{if } o \text{ matches }
        \texttt{\^{}<think>.*?</think>\textbackslash n<answer>.*?</answer>\$},\\
        &\text{and }
        N_{\texttt{<answer>}}(o)=1,\quad
        N_{\texttt{</answer>}}(o)=1,
        \end{aligned}\\
        0, & \text{otherwise}.
    \end{cases}
    \]

Table~\ref{tab:qwen_reward_settings} does not report a separate format compliance metric.

Code training uses Qwen2.5-7B-Instruct on Eurus-2-RL for three epochs.
The evaluation benchmarks are APPS, CodeContests, Codeforces, and TACO.
For a coding query $q$, let $m(q)>0$ denote the integer number of test cases and let $m_{\mathrm{pass}}(q,o)$ denote the number passed by a generated program $o$.
The rewards are
\begin{align*}
\mathcal R_{\mathrm{pass}}(q,o)
&\triangleq
\frac{m_{\mathrm{pass}}(q,o)}{m(q)},\\
\mathcal R_{\mathrm{exec}}(q,o)
&\triangleq
\mathbf 1\{o\text{ compiles and executes without errors on the tests for }q\}.
\end{align*}
Code evaluation uses $16$ responses per problem, a nucleus sampling threshold of $0.95$, and a maximum generation length of $2048$ tokens.
The pass metric averages test case pass rates, and the bug metric is the fraction of programs with compilation or runtime errors.

\subsection{Baseline Implementations}

All additional mathematical comparisons use implementations in the same codebase.
GRPO standardizes a scalarized reward within each rollout group, but its reported adaptive configuration additionally contains the format reward and reward KL term specified above.
GDPO independently standardizes each objective, aggregates the resulting advantages, and applies batch standardization.
SAW uses the implementation variant that weights each independently standardized objective advantage by a coefficient of variation based weight and then applies batch standardization.
DVAO uses a shared denominator formed from objective reward standard deviations.
GD$^2$PO uses a hard mask for conflicting advantages.
Focal Reward is adapted to pointwise rewards rather than reproducing its original pairwise reward construction.
Each trained configuration in Table~\ref{tab:adaptive_rewards} uses three independent training runs.
The external evaluation in Table~\ref{tab:controlled_adaptive} includes two runs for GRPO, GD$^2$PO, and each SA-MRPO configuration, and three runs for the remaining trained configurations.

\section{Additional Results}
\label{app:additional_results}

\begin{figure}[t]
    \centering
    \begin{minipage}[t]{0.49\columnwidth}
        \vspace{0pt}
        \centering
        \includegraphics[width=\linewidth]{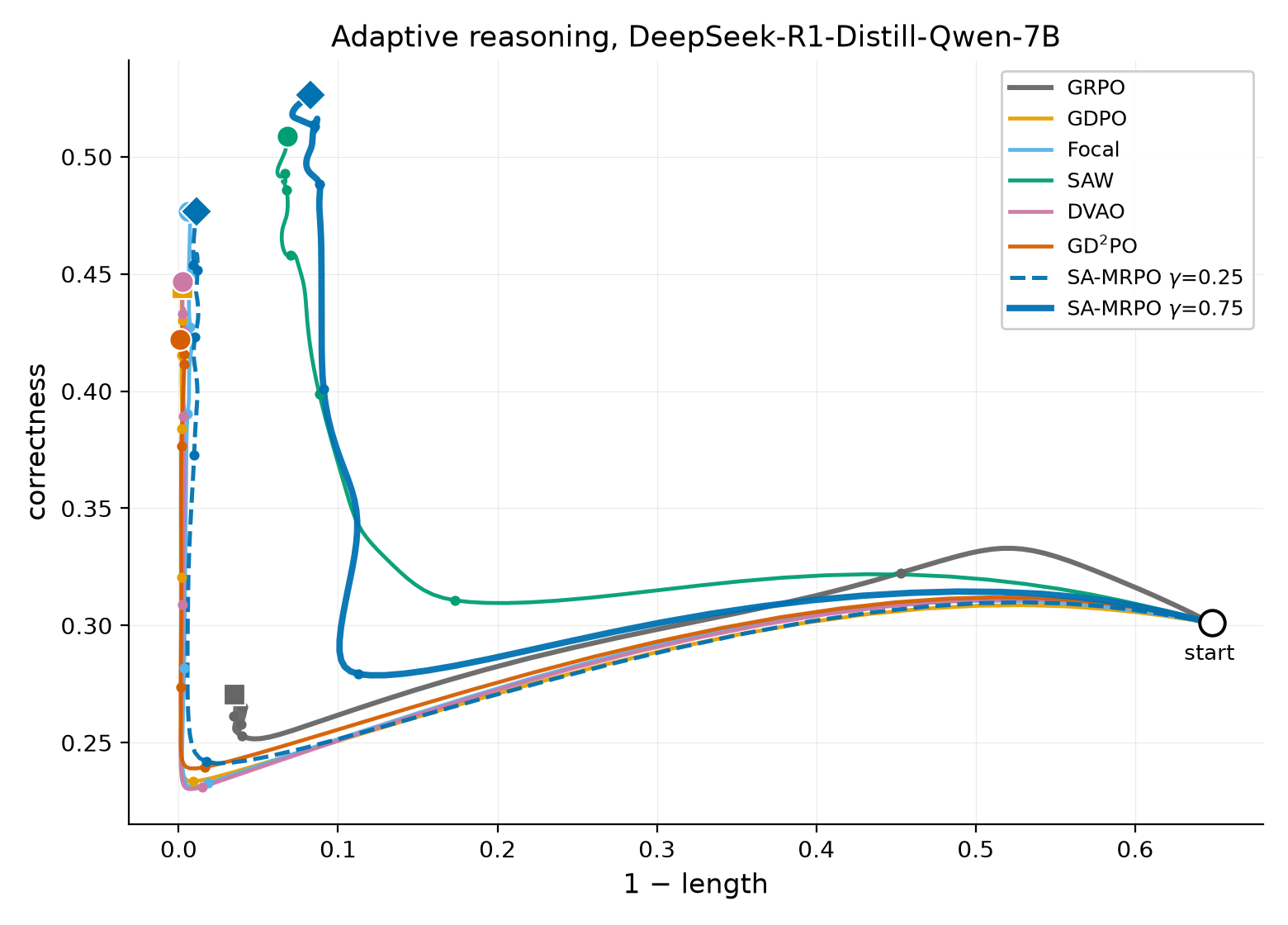}
        \caption{Reward trajectories for adaptive reasoning with DeepSeek-R1-Distill-Qwen-7B. The horizontal axis is one minus the mean length reward, and the vertical axis is mean correctness.}
        \label{fig:reward_trajectories}
    \end{minipage}
    \hfill
    \begin{minipage}[t]{0.49\columnwidth}
        \vspace{0pt}
        \centering
        \includegraphics[width=\linewidth]{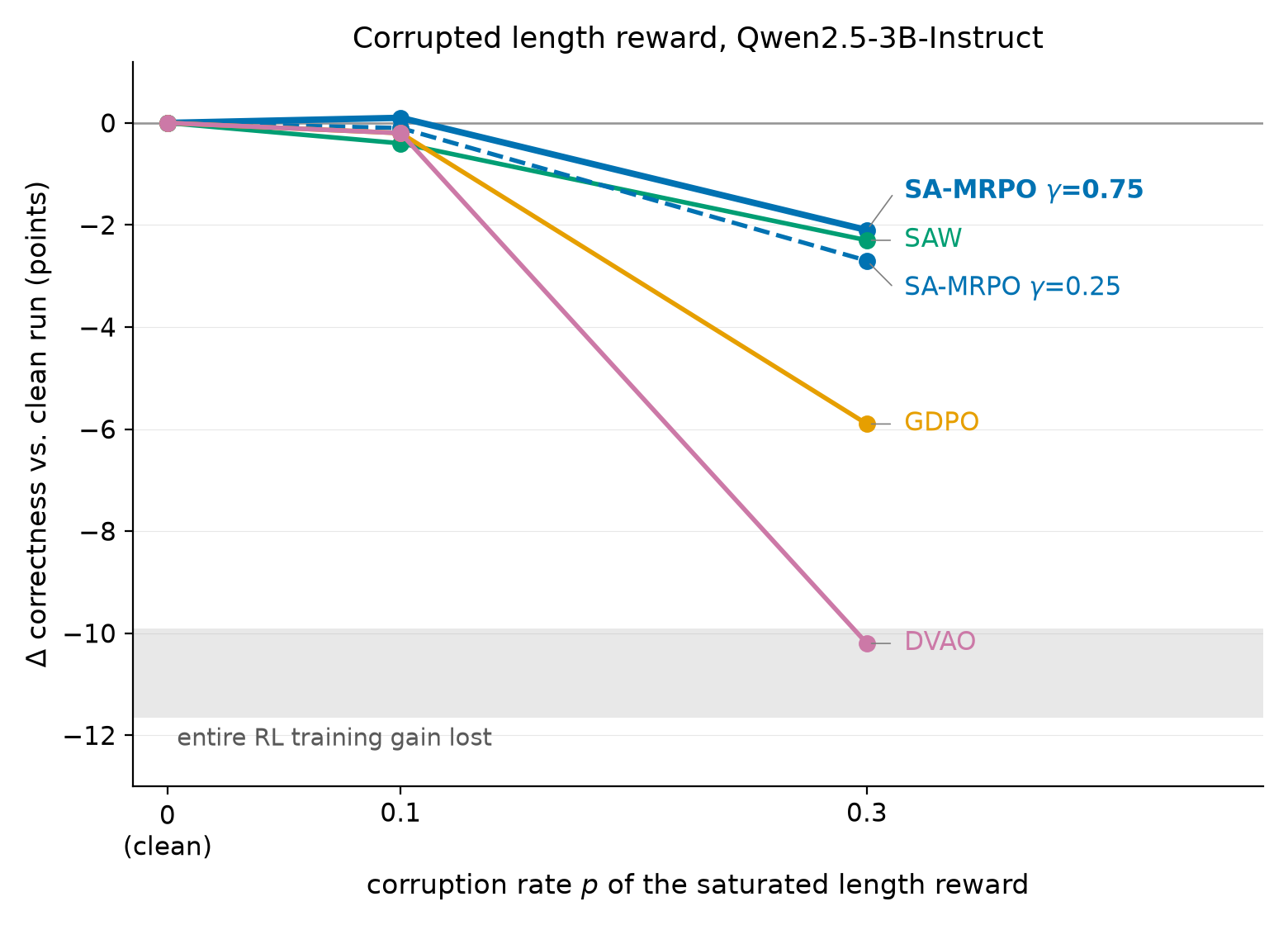}
        \caption{Changes in DeepScaleR correctness relative to clear runs, after training with corrupted length rewards. The shaded band marks the loss of the entire RL training gain in this setting}
        \label{fig:noise}
    \end{minipage}
\end{figure}

\subsection{Adaptive Reward Measurements and Training Diagnostics}
\label{app:adaptive_results}
\label{app:adaptive_diagnostics}

Table~\ref{tab:adaptive_rewards} complements the external benchmark results in Table~\ref{tab:controlled_adaptive} by reporting the correctness--length reward tradeoff on DeepScaleR, with trained configurations averaged over three seeds.
The reported mean correctness increases with $\gamma$, while the mean length reward decreases.
This pattern is consistent with reducing the relative emphasis on the length objective as its reward approaches the upper bound.

\begin{table}[t]
\centering
\small
\caption{Adaptive DeepScaleR evaluation rewards.
Trained configurations use three training seeds, with sample standard deviations of correctness.
Length entries are means.
Base denotes the initial policy measurements.}
\label{tab:adaptive_rewards}
\begin{tabular}{lcc}
\toprule
Method & Correctness reward& Length reward\\
\midrule
Base & $0.327\pm0.004$ & $0.419$ \\
GRPO & $0.296\pm0.004$ & $0.973$ \\
GDPO & $0.448\pm0.005$ & $0.997$ \\
GD$^2$PO & $0.430\pm0.007$ & $0.996$ \\
DVAO & $0.452\pm0.009$ & $0.996$ \\
Focal adaptation & $0.474\pm0.027$ & $0.989$ \\
SAW & $0.510\pm0.006$ & $0.927$ \\
SA-MRPO, $\gamma=0.25$ & $0.476\pm0.004$ & $0.986$ \\
SA-MRPO, $\gamma=0.5$ & $0.485\pm0.015$ & $0.966$ \\
SA-MRPO, $\gamma=0.75$ & $0.533\pm0.008$ & $0.923$ \\
SA-MRPO, $\gamma=1$ & $0.540\pm0.008$ & $0.899$ \\
\bottomrule
\end{tabular}
\end{table}

\begin{figure}[t]
\centering
\includegraphics[width=\linewidth]{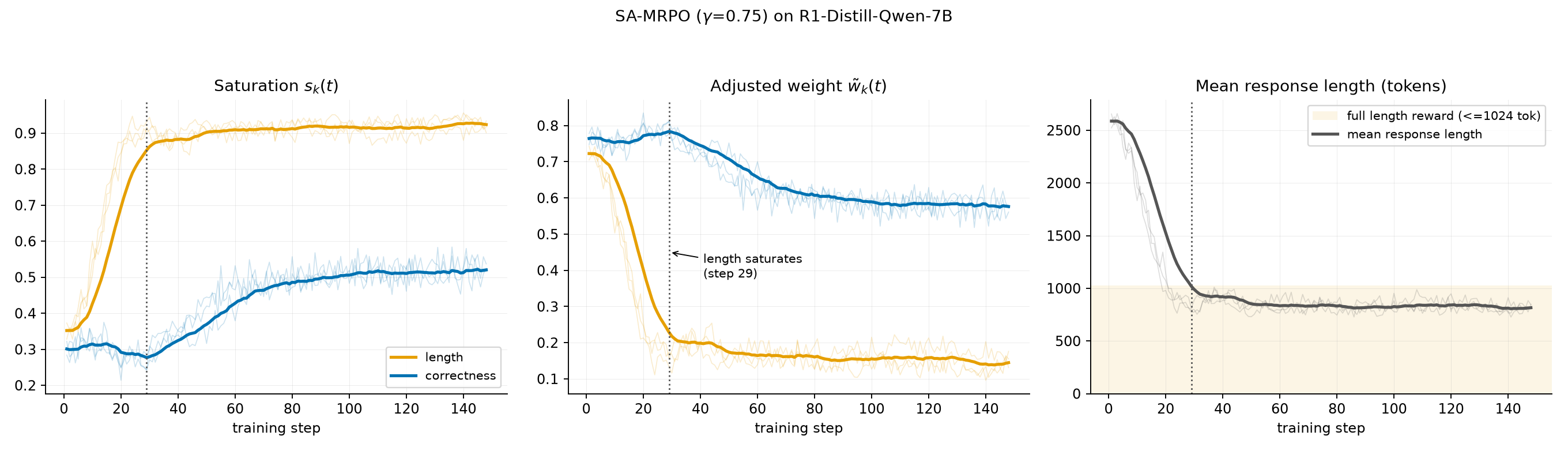}
\caption{Adaptive training with SA-MRPO at $\gamma=0.75$ over three seeds.
Panels show reward means, adjusted objective weights, and mean response length.
}
\label{fig:saturation_mechanism}
\end{figure}

Figure~\ref{fig:saturation_mechanism} tracks the decline in the length coefficient as the mean length reward approaches its upper bound during training with $\gamma=0.75$.
The correctness coefficient remains larger during the later part of training.
The changing coefficient ratio shows that the aggregate advantage places less relative weight on the standardized length advantage.
These coefficients describe advantage aggregation before final batch normalization.

\subsection{Corrupted Length Rewards}
\label{app:corrupted_rewards}

We examine sensitivity to length-reward corruption using Qwen2.5-3B-Instruct, with correctness and the quadratic length reward defined by $(L_{\min},L_{\max})=(1000,4000)$.
A fixed hash of each generated response determines whether its length reward is set to zero, at nominal corruption rates $p=0.1$ and $p=0.3$.
For a fixed rollout batch, zeroing positive length rewards increases measured length headroom and can increase reward variance without changing any response length.
Thus, corruption can affect both headroom-based and dispersion-based weighting rules without any change in the underlying response lengths.
Each corrupted point in Figure~\ref{fig:noise} represents a single training run and reports the change in correctness relative to an uncorrupted run using the same training seed.
The uncorrupted runs improve correctness by $9.9$--$11.7$ percentage points across methods over the base reference of $21.35\%$, averaged over $18$ evaluations before training.
The shaded band marks the corresponding losses that would erase these training gains and return performance to the base-model level.

At $p=0.1$, correctness decreases by $0.2$, $0.4$, $0.2$, and $0.1$ percentage points for GDPO, SAW, DVAO, and SA-MRPO with $\gamma=0.25$, respectively, while SA-MRPO with $\gamma=0.75$ gains $0.1$ percentage points.
At $p=0.3$, the decreases are $5.9$, $2.3$, $10.2$, $2.7$, and $2.1$ percentage points for GDPO, SAW, DVAO, SA-MRPO with $\gamma=0.25$, and SA-MRPO with $\gamma=0.75$, respectively.
DVAO loses approximately its entire gain from RL training, returning close to the base-model level.
Both SA-MRPO settings retain most of their training gains and show smaller losses than GDPO and DVAO in these runs.

\subsection{Saturation Exponent Sweeps}
\label{app:gamma_sweeps}

\begin{figure}[t]
\centering
\includegraphics[width=\linewidth]{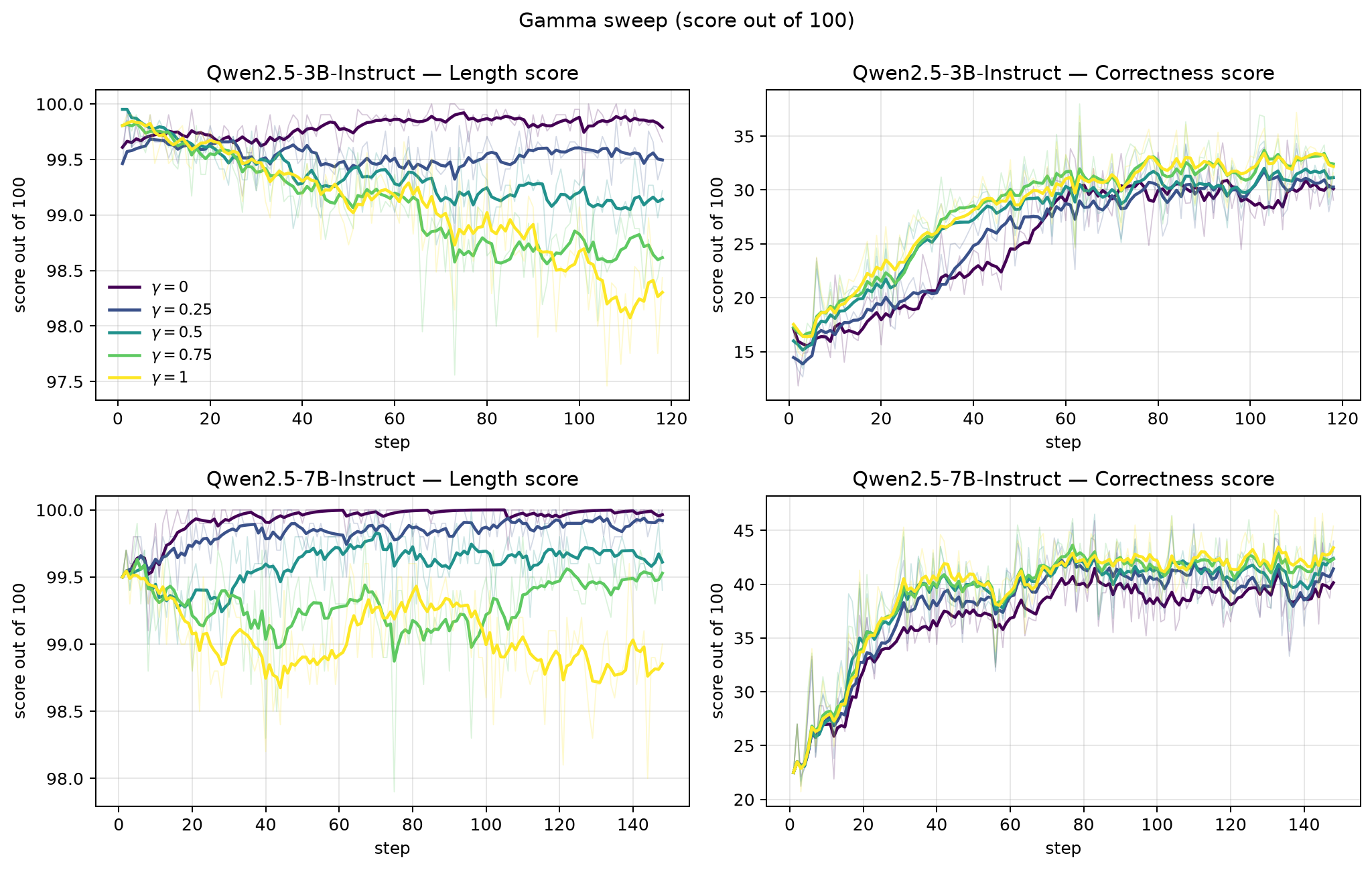}
\caption{Training reward trajectories for Qwen2.5 3B and 7B under different saturation exponents.
Each configuration uses one training run.
Scores are displayed on a scale from zero to one hundred.}
\label{fig:reward_gamma_sweep}
\end{figure}

The sampled exponent study trains Qwen2.5-3B-Instruct and Qwen2.5-7B-Instruct for one epoch with $G=4$, a batch size of $256$, and a maximum response length of $4096$ tokens.
It uses correctness and quadratic length rewards.
Across the training runs in Figure~\ref{fig:reward_gamma_sweep}, larger $\gamma$ is generally associated with lower length rewards.
The correctness trajectories generally improve over $\gamma=0$, but their ordering does not follow $\gamma$ monotonically.
Table~\ref{tab:sagdpo_gamma_sweep_3b} evaluates the resulting 3B policies on the five mathematical benchmarks.
The $\gamma=0.25$ column corresponds to the 3B two objective SA-MRPO configuration in Table~\ref{tab:qwen_reward_settings}.
Among the evaluated settings, $\gamma=0.5$ has the highest accuracy when the five benchmarks receive equal weight.
Increasing $\gamma$ beyond $0.5$ raises the mean exceedance rate without improving mean accuracy in this experiment.

\begin{table}[t]
\centering
\small
\setlength{\tabcolsep}{5pt}
\caption{Sampled 3B exponent study.
Cells show accuracy / responses exceeding $4000$ tokens, both in percent.
Evaluation uses $16$ responses per problem and one training run per configuration.}
\label{tab:sagdpo_gamma_sweep_3b}
\begin{tabular}{lcccccc}
\toprule
Benchmark & Base & $\gamma=0$ & $\gamma=0.25$ & $\gamma=0.5$ & $\gamma=0.75$ & $\gamma=1$ \\
\midrule
AIME24 & 0.6 / 6.2 & 5.0 / 0.0 & 8.5 / 0.6 & 9.0 / 0.7 & 8.7 / 0.8 & 7.4 / 1.1 \\
Minerva & 6.7 / 0.4 & 16.2 / 0.1 & 16.6 / 0.1 & 16.9 / 0.2 & 17.0 / 0.1 & 16.8 / 0.3 \\
AMC23 & 10.7 / 1.7 & 33.2 / 0.0 & 34.9 / 0.1 & 35.6 / 0.2 & 35.3 / 0.2 & 34.8 / 0.4 \\
MATH500 & 26.3 / 0.6 & 57.1 / 0.0 & 58.2 / 0.1 & 58.6 / 0.1 & 58.8 / 0.2 & 58.5 / 0.3 \\
Olympiad & 4.5 / 3.3 & 20.6 / 0.1 & 19.3 / 0.3 & 20.1 / 0.1 & 19.4 / 0.7 & 20.7 / 0.3 \\
\bottomrule
\end{tabular}
\end{table}

\end{document}